# Improving problem solving by exploiting the concept of symmetry

M. A. El-Dosuky[1], M. Z. Rashad[1], T. T. Hamza[1], and A.H. EL-Bassiouny[2]

[1] Department of Computer Sciences, Faculty of Computers and Information sciences, Mansoura University, Egypt
mouh_sal_010@mans.edu.eg
magdi_12003@yahoo.com
Taher_Hamza@yahoo.com

[2] Department of Mathematics, Faculty of Sciences, Mansoura University, Egypt
el_bassiouny@mans.edu.eg

**Abstract**

*This paper investigates the concept of symmetry and its role in problem solving. It first defines precisely the elements that constitute a "problem" and its "solution," and gives several examples to illustrate these definitions. Given precise definitions of problems, it is relatively straightforward to construct a search process for finding solutions. Finally this paper attempts to exploit the concept of symmetry in improving problem solving.*

## 1  Introduction

This section investigates the concept of symmetry and its role in problem solving. This section also defines precisely the elements that constitute a "problem" and its "solution," and gives several examples to illustrate these definitions. Given precise definitions of problems, it is relatively straightforward to construct a search process for finding solutions.

### 1.1 Symmetry

Symmetry is a fundamental part of geometry, shapes, and nature in general. It creates patterns that aid us organize our world conceptually [1].

Many researchers ([1], [2], and [3]) support that symmetry is all around us and even though it does not seem to be mathematical. However, it appears that there is little research concerning the development of symmetry in the field of artificial intelligence, especially problem solving. According to Leikin, Berman and Zaslavsky, symmetry has a special role in problem solving [4]. In their studies with secondary mathematics teachers, they emphasize that symmetry connects various branches of mathematics such as algebra, geometry, probability, and calculus, and present it as a useful problem-solving tool. Symmetry has been exploited for automated reasoning in geometry theorems with Prolog [5].

### 1.2 Problem and solution

A ***problem*** can be defined by five components[6]:
1. The ***initial state*** that the agent starts in.
2. The ***state space*** is the set of all states reachable from the initial state.
3. Possible ***actions***, commonly defined using a successor function.
4. The **goal test**, which determines whether a given state is a goal state.
5. A **path cost** is a function that assigns a numeric cost to each path.

A ***solution*** to a problem is a path from the initial state to a goal state. A ***path*** in the state space is a sequence of states connected by a sequence of actions. **Solution quality** is measured by path cost function. The **optimal solution** has the lowest path cost among all solutions. The **step cost** of taking action *a* to go from state *x* to state *y* is denoted c(*x*, *a*, *y*)[6].

***Problem formulation*** is the process of deciding what actions and states to consider, and follows goal formulation. In the following we are formulating some simple problems[6].

*Example 1: Problem formulation of Vacuum Cleaner World*

The world has two locations, namely square A and square B. The vacuum cleaner can perceive its location and if there is dirt. It can choose to move left, move right, suck up dirt. A simple agent function can be used, for example, if current square is dirty then suck, else move to the other square.

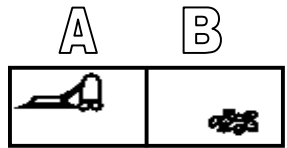

Fig. 1 the vacuum cleaner world

The formulation:
- **States**: the agent can be in one of two locations; each might or might not contain dirt. For *n* locations, we have $n \times 2^n$ states. Since we have 2 locations, then we have $2 \times 2^2 = 8$ possible states, as shown in fig. 2.
- **Initial state**: any state can be considered an initial state.
- **Successor function**: generates the legal states that result from trying three actions (Left, Right, and Suck)
- **Goal test**: checks whether all the squares are clean.
- **Path cost**: number of steps in path, or electrical power to run the vacuum cleaner.

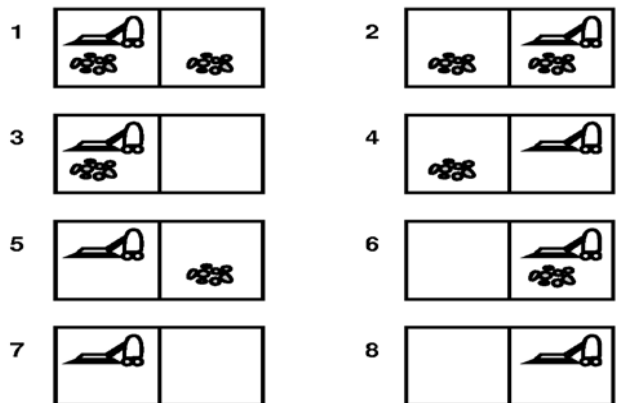

Fig. 2 the vacuum cleaner state space

### Example 2: Missionaries and Cannibals problem

There are three cannibals and three missionaries in one side of the lack. Please help the three cannibals and three missionaries to move the other side of the lack with a boat which can take at maximum two entities. Notice that when there is on one side more cannibals than missionaries the cannibals eat Missionaries[7].

The formulation:
- **state**: (number of missionaries in right side, number of cannibals in right side, side of lack) where side of lack can be : right =1 or left = 0
- **Initial state** : (3,3,1)
- **Goal state** : (0,0,0)
- **Path cost** : Unit for each step
- **Successor function** :

| Action | Move left | Move right |
|---|---|---|
| One missionary | – (1,0,1) | +(1,0,1) |
| One cannibal | – (0,1,1) | +(0,1,1) |
| Two missionary | – (2,0,1) | +(2,0,1) |
| Two cannibal | – (0,2,1) | +(0,2,1) |
| 1 mis., 1 cannibal | – (1,1,1) | +(1,1,1) |

### Example 3: Towers of Hanoi problem

There are three towers A, B, C. We found three different type of vicious circle (small, medium, large) in order from above in tower A[8].
We want to move the three vicious circles from A to C but it must be in the same order in A and you can use tower B as bridge. You can't put larger circle above a smaller one. Don't put large above medium or small. Don't put medium above small.

The formulation:
- **States** :
  State (number of circles, tower to move from, tower to move to)
  Where

1 for larger one
2 for medium one
3 for smallest one

- **Initial state** : All circles in tower A
- **Goal state** : All circles in tower C
- **Path cost** : Unit for each step
- **Successor function :**

|        | Small   | medium  | Larger  |
|--------|---------|---------|---------|
| A to B | (3,A,B) | (2,A,B) | (1,A,B) |
| A to C | (3,A,C) | (2,A,C) | (1,A,C) |
| B to A | (3,B,A) | (2,B,A) | (1,B,A) |
| B to C | (3,B,C) | (2,B,C) | (1,B,C) |
| C to A | (3,C,A) | (2,C,A) | (1,C,A) |
| C to B | (3,C,B) | (2,C,B) | (1,C,B) |

## 1.3 Problems based on knowledge of states

Problems can be classified based upon knowledge of states and actions into four categories[6]:

- Single-state problem
- Multiple-state problem
- Contingency problem
- Exploration problem

In a ***single-state problem***, the agent has complete world state knowledge and complete action knowledge. The agent always knows its world state. For a matter of simplicity and focusing on the symmetry concept, we shall restrict the study of this paper to this kind of problems.

On the other hand, in a ***multiple-state problem***, the agent has incomplete world state knowledge and incomplete action knowledge. The agent only knows which group of world states it is in. In the ***contingency problem***, it is impossible to define a complete sequence of actions that constitute a solution in advance because information about the intermediary states is unknown. In an ***exploration problem***, the state space and effects of actions are unknown. This is kind of problems is considered very difficult.

Let us consider the vacuum cleaner problem as a one-state problem. If the environment is completely accessible, the vacuum cleaner always knows where it is and where the dirt is. The solution then is reduced to searching for a path from the initial state to the goal state. This is shown in fig 3.

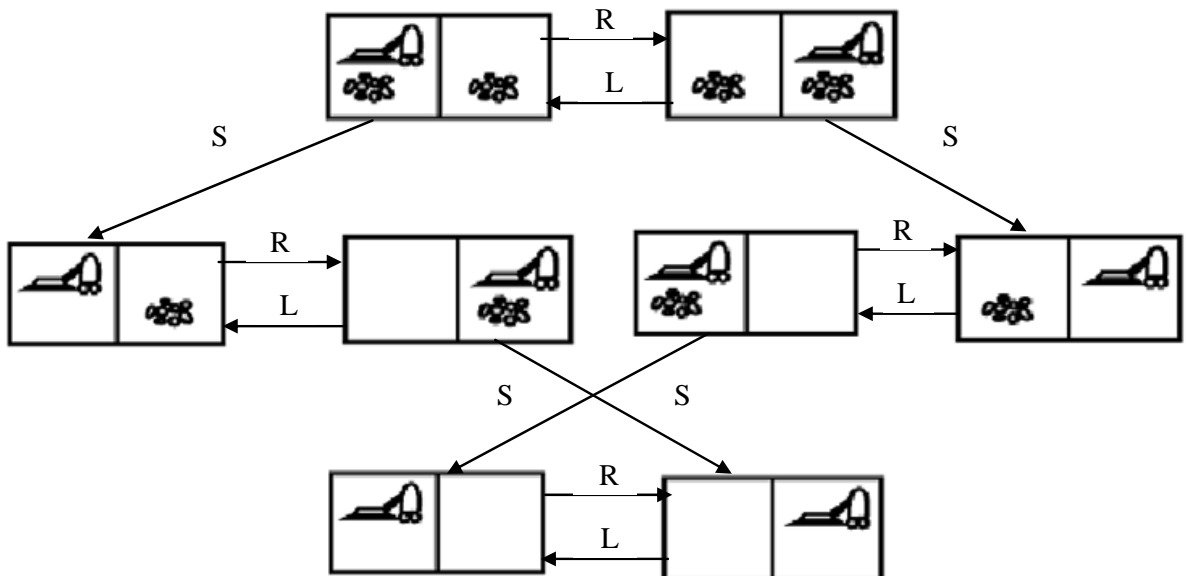

Fig. 3 States for the search: The world states 1-8.

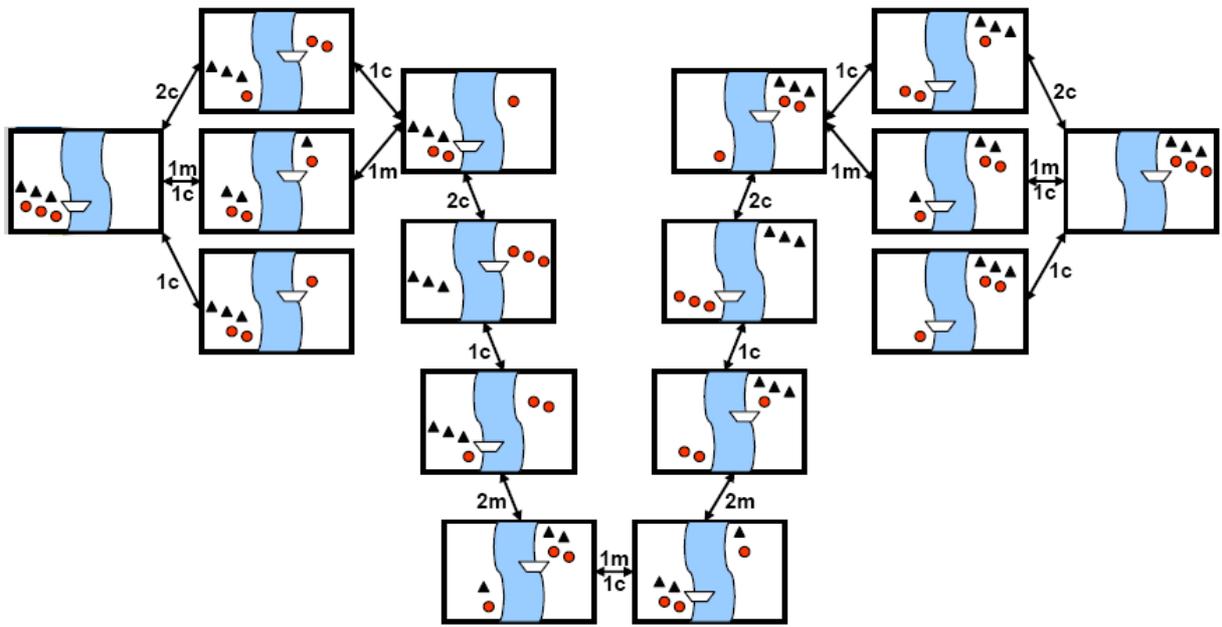

Fig. 4 State space of missionaries and cannibals, 16 states.

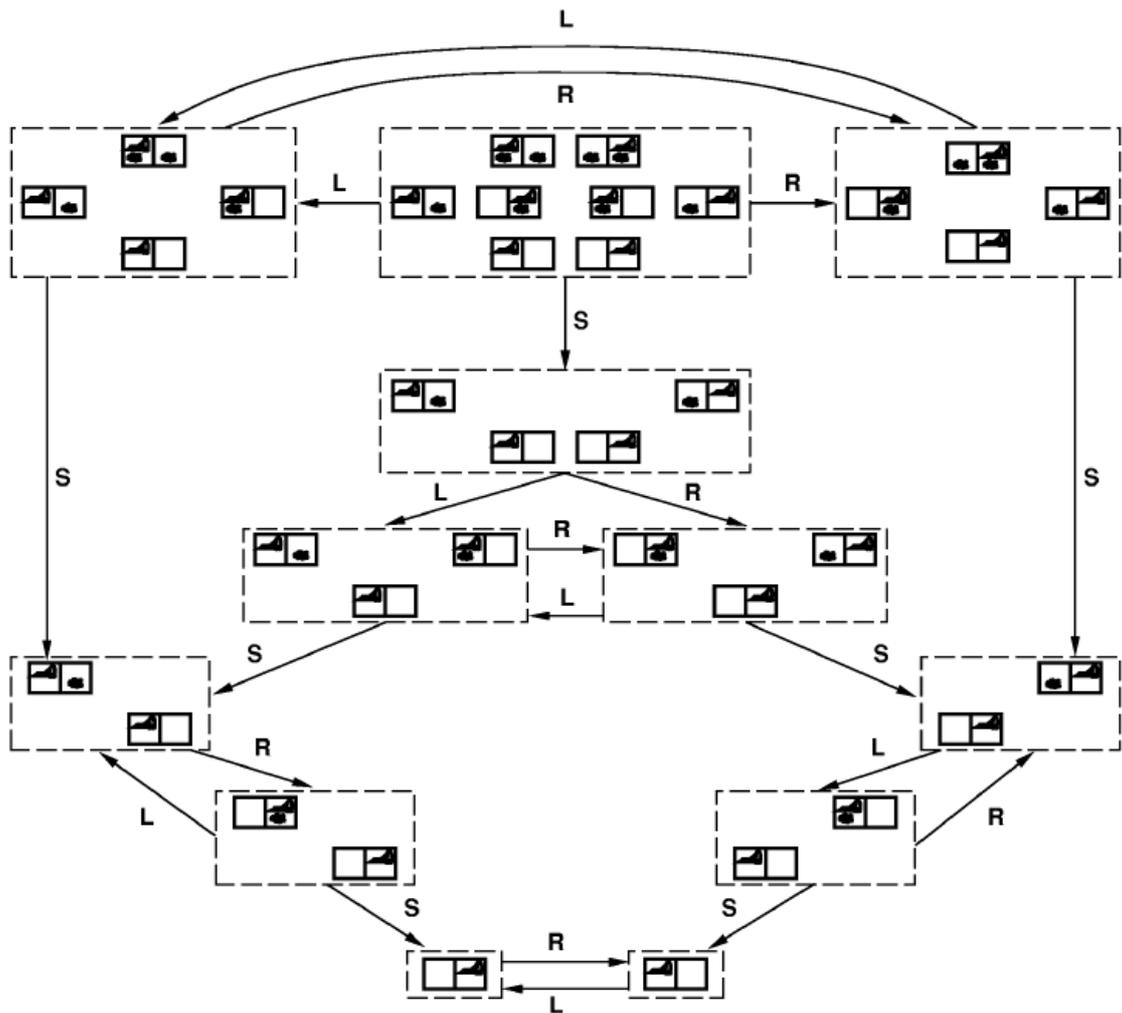

Fig. 5 States for the search: The power set of the world states.

Similarly, let us consider the missionaries and cannibals problem as a one-state problem. If the environment is completely accessible, the solution then is reduced to searching for a path from the initial state to the goal state. This is shown in fig 4.

Although this is considered outside the focus of this study, we shall give an example of the other case of multiple-state to see how the state space will be complicated. Let us consider the vacuum cleaner world as a multiple-state problem. If it has no sensors, it doesn't know where it or the dirt is. It can still solve the problem. Here, states are knowledge states. This is shown in fig 5.

### 1.4 problem-solving agents
A problem-solving agent is a kind of goal-based agent. It decides what to do by finding sequences of actions that lead to desirable states. Here is the general program of the problem solving agent:

> **function** PROBLEM-SOLVING-AGENT(*percept*) **returns** an action
> **inputs:** *percept, a* percept
> **static:** *solution,* an action sequence, initially empty
>       *state,* description of the current world state
>       *goal,* a goal, initially null
>       *problem,* a problem formulation
>
> *state* ← UPDATE-STATE(*state, percept*)
>
> **if** *solution* is empty **then do**
> *goal* ← FORMULATE-GOAL(*state*)
> *problem* ← FORMULATE-PROBLEM(*state, goal*)
> *solution* ← SEARCH( *problem*)
> *action* ← FIRST(*solution*)
> **return** *action*

Fig. 6 problem-solving agent

## 2. Exploiting Symmetry
This section studies how to exploit the concept of symmetry in problem solving.

### 2.1 vacuum cleaner revisited
Let us reconsider the vacuum cleaner problem as a one-state problem as shown in figure 3.

Clearly symmetry is obvious in the state space as seen also in state space of missionaries and cannibals.

However for more clarification let us utilize another simplified formulation. The state now is
                      Sign(left, right)
Where sign determine the location of the cleaner : - is left and + for right. And left, right represent the existence of dirt in left, right respectively.

The new state space is shown in figure 7. And a combined state space is shown in fig. 8.

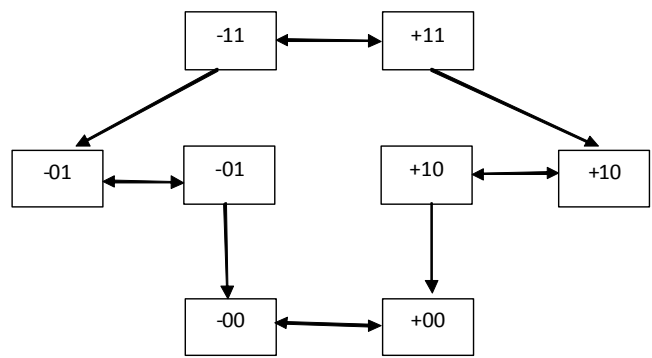

Fig. 7  States for the search using simplified formulation.

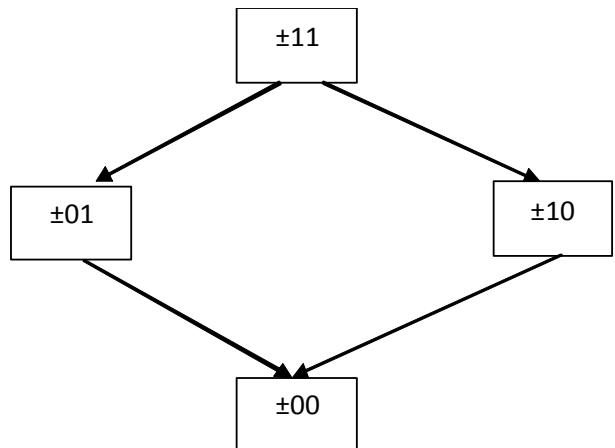

Fig. 8 combined states for the search using simplified formulation

## 3. Conclusion

By comparing fig 7 and fig 8, from the vacuum state space we noticed how the state space is simplified exponentially using symmetry.

The problem solving agent can improve operation if it utilizes one of the following strategies:
- bidirectional search ,
- forward search from initial while regarding the mirror state of the goal, or
- Backward search from goal while regarding the mirror state of the initial.